\documentclass{article} 
\usepackage{iclr2022_conference,times}

\usepackage{amsmath,amsfonts,bm}

\def\eqref#1{(\ref{#1})}

\def\1{\bm{1}}

\DeclareMathAlphabet{\mathsfit}{\encodingdefault}{\sfdefault}{m}{sl}
\SetMathAlphabet{\mathsfit}{bold}{\encodingdefault}{\sfdefault}{bx}{n}

\DeclareMathOperator*{\argmax}{arg\,max}

\usepackage{hyperref}
\usepackage{url}
\usepackage{xcolor,colortbl}
\usepackage{color}
\definecolor{Gray}{gray}{0.9}

\usepackage[utf8]{inputenc} 
\usepackage[T1]{fontenc}    
\usepackage{hyperref}       
\usepackage{url}            
\usepackage{booktabs}       
\usepackage{amsfonts}       
\usepackage{nicefrac}       
\usepackage{microtype}      

 \usepackage{graphicx}
\usepackage{subfigure} 
\usepackage{pifont}

\usepackage{makecell}

\usepackage{adjustbox}
\usepackage{lipsum}
\usepackage{color}
\usepackage{wrapfig}
\usepackage{booktabs}
\usepackage{multirow,mathtools} 
\usepackage{algorithm,algpseudocode}
\usepackage{threeparttable}
  \usepackage{xcolor}

 \algnewcommand{\algorithmicforeach}{\textbf{for each}}
\algdef{SE}[FOR]{ForEach}{EndForEach}[1]
  {\algorithmicforeach\ #1\ \algorithmicdo}
  {\algorithmicend\ \algorithmicforeach}

\definecolor{lightblue}{rgb}{0.68, 0.85, 0.9}
\definecolor{lightgreen}{rgb}{0.56, 0.93, 0.56}
\definecolor{lightskyblue}{rgb}{0.53, 0.81, 0.98}
\definecolor{non-photoblue}{rgb}{0.64, 0.87, 0.93}
\definecolor{magicmint}{rgb}{0.67, 0.94, 0.82}
\definecolor{mossgreen}{rgb}{0.68, 0.87, 0.68}
\definecolor{salmon}{rgb}{1.0, 0.55, 0.41}
\definecolor{babypink}{rgb}{0.96, 0.76, 0.76}

\DeclareMathOperator*{\minimize}{\text{minimize}}

\DeclareMathAlphabet\mathbfcal{OMS}{cmsy}{b}{n}
\newcommand{\Def}[0]{\mathrel{\mathop:}=}

 \newcommand{\SL}[1]{\textcolor{red}{SL: #1}}

 \def\remark{\addtocounter{remark}{1}\def\@currentlabel{\theremark}%
\emph{Remark~\theremark}. } \makeatother

\newcounter{remark}

\def\btheta{\boldsymbol{\theta}}
\def\bdelta{\boldsymbol{\delta}}

\def\bdelta{\boldsymbol{\delta}}

\usepackage{color, colortbl}
\definecolor{Gray}{gray}{0.9}
\definecolor{Orange}{rgb}{1,0.5,0}

\makeatletter
\newcommand*{\rom}[1]{\expandafter\@slowromancap\romannumeral #1@}
\makeatother
\newcommand{\mycomment}[1]{}

\newcommand{\DS}{{DS}}
\newcommand{\RS}{{RS}}
\newcommand{\FO}{{FO}}
\newcommand{\ZO}{{ZO}}
\newcommand{\ZOAEDS}{{ZO-AE-DS}}
\newcommand{\FOAEDS}{{FO-AE-DS}}
\newcommand{\FODS}{{FO-DS}}
\newcommand{\ZODS}{{ZO-DS}}
\newcommand{\RGE}{{RGE}}
\newcommand{\CGE}{{CGE}}

\newcommand{\AEnc}{{AE}}

\usepackage{accents}
\newlength{\dhatheight}

\iclrfinalcopy

\title{How to Robustify Black-Box ML Models? 
\\A Zeroth-Order Optimization Perspective}

\author{
  Yimeng Zhang$^1$, 
  Yuguang Yao$^{1}$, 
  Jinghan Jia$^{1}$,
  Jinfeng Yi$^{2}$, 
  Mingyi Hong$^{3}$, 
  Shiyu Chang$^{4}$, 
  Sijia Liu$^{1,5}$
 \\
  $^{1}$ Michigan State University, 
  $^{2}$ JD AI Research,
  $^{3}$ University of Minnesota,
  \\
  $^{4}$ UC Santa Barbara,
  $^{5}$ MIT-IBM Watson AI Lab, IBM Research
}

\begin{document}

\maketitle

\begin{abstract}

The lack of adversarial robustness has been  recognized as an important issue for state-of-the-art machine learning (ML) models, e.g., deep neural networks (DNNs). Thereby, robustifying ML models against adversarial attacks is now a major focus of research. However, nearly all existing defense methods, particularly for robust training,  made the \textit{white-box} assumption that the defender has the access to the details of an ML model (or its surrogate alternatives if available), e.g., its architectures and parameters. Beyond existing works, in this paper we aim to address the problem of \textit{black-box defense}: {How to robustify a black-box model using just input queries and output feedback?}
Such a problem   arises in practical scenarios, where the owner of the predictive model is reluctant to share model information in order to preserve  privacy. To this end, we propose a general notion of defensive operation that can be applied to black-box models,  and design it through the lens of denoised smoothing ({\DS}), a first-order (FO) certified defense technique. To allow  the design of merely using model queries, we further integrate {\DS} with the zeroth-order (gradient-free) optimization. However, a direct implementation   of  zeroth-order (ZO) optimization suffers a high variance of gradient estimates, and thus leads to ineffective defense. To tackle this problem, we next propose to prepend an autoencoder ({\AEnc})  to a given (black-box) model so that DS can be trained using variance-reduced ZO optimization. We term the eventual defense as ZO-{\AEnc}-{\DS}. In practice, we empirically show that ZO-{\AEnc}-{\DS} can achieve improved accuracy, certified robustness, and  query complexity over existing baselines. And the effectiveness of our approach is justified under both image classification and image reconstruction tasks. Codes are available at \href{https://github.com/damon-demon/Black-Box-Defense}{\texttt{https://github.com/damon-demon/Black-Box-Defense}}. 

\end{abstract}

\section{Introduction}

ML models, DNNs in particular, have achieved   remarkable   success  owing to their superior predictive performance. However, they often lack robustness. For example, imperceptible but carefully-crafted input perturbations can fool the decision of a well-trained ML model. 
These input perturbations refer to \textit{adversarial perturbations}, and the adversarially perturbed (test-time) examples are known as \textit{adversarial examples} or \textit{adversarial attacks} \citep{Goodfellow2015explaining,carlini2017towards,papernot2016limitations}. Existing studies have shown that 
it is not difficult to generate adversarial attacks. Numerous
  attack generation methods     have   been designed and successfully applied to \textit{(i)} different use cases from the digital world to the physical world, \textit{e.g.}, image classification \citep{brown2017adversarial,li2019adversarial,xu2019structured, yuan2021meta}, object detection/tracking \citep{eykholt2017robust,xu2020adversarial,sun2020towards}, and image reconstruction \citep{antun2020instabilities,raj2020improving, vasiljevic2021self}, and \textit{(ii)} different types of victim models, \textit{e.g.}, white-box models whose details can be accessed by adversaries \citep{madry2017towards, carlini2017towards,tramer2020adaptive,croce2020reliable, wang2021di}, and black-box models whose information is not disclosed to adversaries \citep{
  papernot2017practical,tu2019autozoom,ilyas2018black, liang2021parallel}.

Given the prevalence of adversarial attacks, methods to robustify ML models  are now a major focus in research.  For example,
adversarial training (AT) \citep{madry2017towards}, which has been poised one of   {the most}  effective defense methods \citep{athalye2018obfuscated}, 
employed
min-max optimization to minimize  the worst-case (maximum)
training loss induced by   adversarial attacks. Extended from AT, various empirical defense methods were proposed, ranging from  supervised learning, semi-supervised learning, to unsupervised learning \citep{madry2017towards,zhang2019theoretically,shafahi2019adversarial,zhang2019you,carmon2019unlabeled,chenliu2020cvpr, zhang2021defense}. In addition to empirical defense, certified defense is   another research focus, which aims to train provably robust ML models and provide  certificates of   robustness \citep{wong2017provable, raghunathan2018certified,katz2017reluplex,salman2019provably, salman2020denoised, salman2021certified}. 
Although exciting progress has been made in adversarial defense, nearly all existing 
works ask
a defender   to perform over \textit{white-box} ML models (assuming non-confidential model architectures and parameters). However, the white-box assumption may restrict the defense application in practice. 
For example, a model owner may refuse      to share the model details, since  disclosing model information could hamper the owner's privacy, \textit{e.g.},   model inversion attacks lead to training data  leakage   \citep{fredrikson2015model}. 
Besides the privacy consideration, 
the white-box defense built upon the (end-to-end) robust training (\textit{e.g.}, AT)  is computationally intensive, and thus is difficult to scale when    robustifying   multiple models. 
For example, 
in the medical domain, there exist     massive pre-trained ML models for different diseases using hundreds of neuroimaging datasets  \citep{sisodiya2020enigma}. Thus, robustly retraining all models becomes impractical.
Taking the model privacy and the defense efficiency 
into consideration, we ask: 
\begin{center}{
\textit{Is it possible to design an adversarial defense over  \textbf{black-box} models using only model queries?}
}
\end{center}

 \begin{wrapfigure}{r}{85mm}
\vspace{-7mm}
\centering{
\begin{tabular}{c}
\hspace*{-6.5mm}
\includegraphics[width=0.65\textwidth]{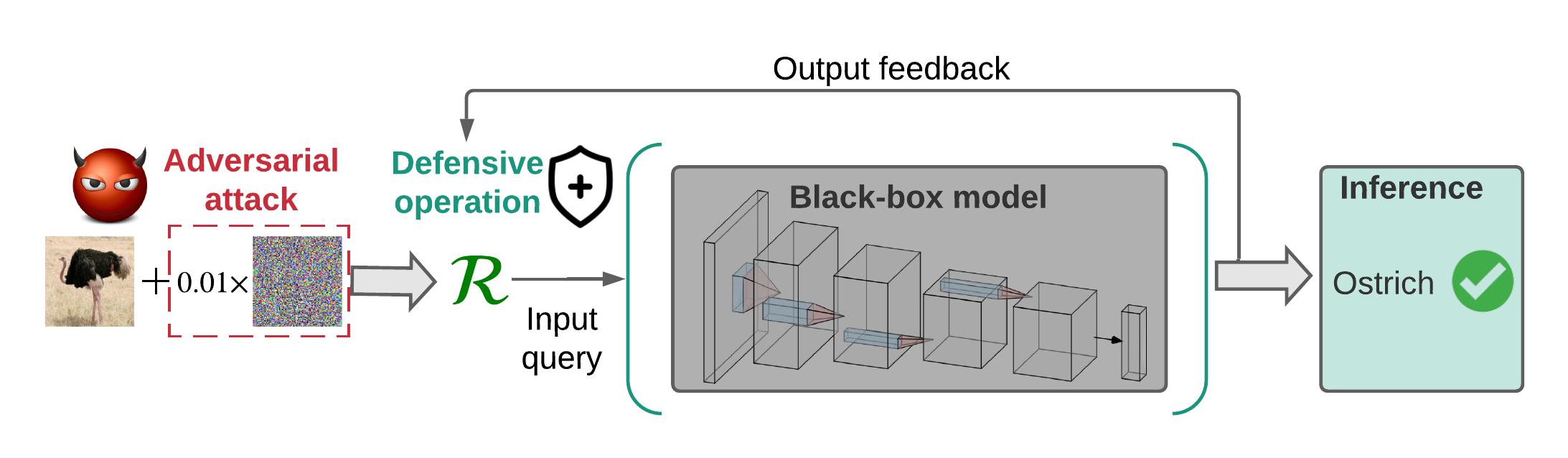}
\end{tabular}
}
\vspace{-7mm}
\caption{\footnotesize{Illustration of defense against adversarial attacks for entirely black-box models.
}}
\vspace{-3mm}
\label{fig: motivation_problem}
\end{wrapfigure}
Extending adversarial defense to the black-box   regime (that we call `black-box defense')  
is highly non-trivial due to
the challenge of black-box optimization  (\textit{i.e.}, learning over black-box models). 
To tackle this problem, the  prior   work   \citep{salman2020denoised}   leveraged  \textit{surrogate models} as approximations of     the black-box models, over which defense can be conducted following the white-box setup. 
Yet, this still requires to have access to the information on the victim model type and its function. In practice, those conditions could be difficult to achieve. For example, if the domain knowledge related to  medicine or healthcare  is lacking   \citep{qayyum2020secure,finlayson2019adversarial}, then it will be   difficult to determine a proper surrogate model of  a medical ML system.
Even if a black-box model estimate can be obtained using the  model inversion technique
\citep{kumar2019modelinversion}, a significantly large number of model queries are needed  even just for tackling a MNIST-level prediction task \citep{oh2019reverseengineerblackbox}. 
Different from  \citep{salman2020denoised}, we study an \textit{authentic black-box scenario}, in which the interaction between defender and model is only based on input-output function queries (see Fig.\,\ref{fig: motivation_problem}).
To our best knowledge, 
this is the first work to tackle the problem of query-based black-box defense.

\paragraph{Contributions.} We summarize our contributions below. 

\noindent \ding{172}
(Formulation-wise) We formulate the problem  of black-box defense and investigate it through the lens of zeroth-order (ZO) optimization.
Different from existing 
works, our paper aims to design the restriction-least black-box defense and our formulation is built upon a query-based black-box setting, which avoids the use of surrogate models. 

\noindent \ding{173}
(Methodology-wise)
We propose a novel black-box defense approach, \underline{ZO}  \underline{A}uto\underline{E}ncoder-based \underline{D}enoised \underline{S}moothing     ({\ZOAEDS}), which  is able to  tackle the challenge of ZO optimization in high dimensions and convert a pre-trained  non-robust ML model into a certifiably robust model using only function queries.

\noindent \ding{174}
(Experiment-wise)
We verify the efficacy of our method through an extensive experimental study. In the task of image classification, the proposed {\ZOAEDS} significantly outperforms the ZO baseline built upon \citep{salman2020denoised}.
For instance, we can  improve the certified robust accuracy of  ResNet-110
 on CIFAR-10 from $19.16 \%$ (using baseline)
to $54.87 \%$ (using {\ZOAEDS})
under adversarial perturbations with $\ell_2$ norm less than $64/255$.
We also empirically   show that   our proposal stays   effective even in the task of image reconstruction.

\section{Related work}
\label{related_work}

\paragraph{Empirical  defense.}

An immense number of  defense methods have been proposed, aiming to improve model robustness against adversarial attacks.
Examples include detecting adversarial attacks~\citep{guo2017countering,meng2017magnet,gong2017adversarial, grosse2017statistical,metzen2017detecting} and training robust ML models \citep{madry2017towards,zhang2019theoretically,shafahi2019adversarial,Wong2020Fast,zhang2019you,athalye2018obfuscated,cheng2017maximum, wong2017provable,salman2019provably, raghunathan2018certified,katz2017reluplex}.
In this paper, we focus on advancing the algorithm foundation of robust training over black-box models. 
 Robust training  can be broadly divided into two categories: empirical defense 
 and certified defense.
 In the former category, the most representative method is AT (adversarial training) that formulates adversarial defense as a two-player game (between attacker and defender) \citep{madry2017towards}. Spurred by   AT,  empirical defense has developed rapidly. For example,  in \citep{zhang2019theoretically}, TRADES was proposed to seek the optimal trade-off between accuracy and robustness.
 In \citep{stanforth2019labels, carmon2019unlabeled},  unlabeled data and self-training  were shown effective   to improve  adversarial defense in both robustness and generalization. 
In \citep{shafahi2019adversarial,Wong2020Fast,zhang2019you,andriushchenko2020understanding}, to improve the scalability of adversarial defense, computationally-light alternatives of AT    were developed. Despite the effectiveness of empirical defense against adversarial attacks \citep{athalye2018obfuscated}, it lacks theoretical guarantee (known as `certificate') for the achieved robustness.
Thus, the problem of certified defense arises.

\paragraph{Certified defense.} 
Certified defense
seeks to provide a provably guarantee of   ML models. 
One line of research focuses on 
post-hoc formal verification of a pre-trained ML model. The certified robustness is then given by a `safe' input perturbation region, within which any perturbed inputs will not fool the given   model \citep{katz2017reluplex,ehlers2017formal,bunel2018unified,dutta2017output}. 
Since the exact verification  
is computationally intensive, 
a series of work \citep{raghunathan2018certified,dvijotham2018dual,wong2017provable,weng2018towards,weng2018evaluating,wong2018scaling} proposed `incomplete' verification, which utilizes convex relaxation to over-approximate the output space of a predictive model when facing input perturbations.
 Such a relaxation leads to fast computation in the verification process but only proves a lower bound of the exact robustness guarantee.
Besides the  post-hoc model verification with respect to each input example, another line of research focuses on in-processing certification-aware training and prediction.  For example, 
randomized smoothing ({\RS}) transforms an empirical classifier into a provably robust one by convolving the former with an isotropic Gaussian distribution. It was shown in \citep{cohen2019certified} that 
{\RS} can  provide formal guarantees for adversarial robustness. 
Different types of   {\RS}-oriented provable defenses have been developed, such as  adversarial smoothing  \citep{salman2019provably}, denoised smoothing \citep{salman2020denoised}, smoothed ViT \citep{salman2021certified}, and 
feature smoothing \citep{addepalli2021boosting}.

\paragraph{Zeroth-order (ZO) optimization for adversarial ML.}

ZO optimization methods are gradient-free counterparts of first-order ({\FO}) optimization methods \citep{liu2020primer}. They approximate the {\FO}  gradients   through function value based gradient estimates. Thus,
{\ZO} optimization is quite useful to solve black-box problems when explicit expressions of their gradients are difficult to compute or infeasible to obtain.
In the area of adversarial ML, ZO optimization has become a principled approach to generate adversarial examples from   black-box victim ML models \citep{chen2017zoo,ilyas2018black, ilyas2018prior, tu2019autozoom, liu2018signsgd, liu2019min, Huang2020transfer,  cai2020zeroth, cai2021zeroth}. Such ZO optimization-based attack generation methods can be as effective as state-of-the-art white-box attacks, despite only having access to the inputs and outputs of the targeted  model.
For example, the work \citep{tu2019autozoom} leveraged  the white-box decoder to map the generated low-dimension perturbations back to the original input dimension. Inspired by \citep{tu2019autozoom}, we leverage the autoencoder architecture to tackle the high-dimension challenge of ZO optimization in black-box defense.
 Despite the widespread application of ZO optimization to black-box attack generation, few work studies the problem of black-box defense.

\section{Problem Formulation: Black-Box Defense}
\label{sec: prob_setup}
In this section, we    formulate the problem of black-box defense, \textit{i.e.},  robustifying black-box ML models without having any model  information such as  architectures and parameters.

\paragraph{Problem statement.}

Let $f_{\boldsymbol \theta_{\mathrm{bb}}}(\mathbf x )$ denote a pre-defined \textit{\underline{b}lack-\underline{b}ox (bb) predictive model}, which can  map an input example $\mathbf x$ to a prediction. In our work,  $f_{\boldsymbol \theta_{\mathrm{bb}}}$ can be either an image classifier or an image reconstructor. 
 For simplicity of notation, we will {drop} the model parameters $\boldsymbol \theta_{\mathrm{bb}}$ when referring to a black-box model. 
 The \textit{threat model}  of our interest is given  by   norm-ball constrained adversarial attacks \citep{Goodfellow2015explaining}.
 To defend against these attacks, existing   approaches  commonly require  the {white-box} assumption  of  $f$ \citep{madry2017towards} or 
   have access to white-box surrogate models of  $f$ \citep{salman2020denoised}. 
   Different from the prior works, we study
 the problem of \textit{black-box defense}   when 
  the owner of $f$  is  not able  to  share  the  model  details.  
  Accordingly, the only mode of interaction with the black-box system
is via submitting inputs and receiving the corresponding
predicted outputs.
The formal statement of {black-box defense} is  given below:
\begin{center}
	\vspace{-2mm}
	\setlength\fboxrule{0.0pt}
	\noindent\fcolorbox{black}[rgb]{0.95,0.95,0.95}{\begin{minipage}{0.98\columnwidth}
				\vspace{-0.07cm}
	\textbf{(Black-box defense)}		Given a black-box base model
   $f$, can we develop a defensive operation $\mathcal R$
 using just input-output \textit{function queries} so as
to produce the robustified  model $\mathcal R(f)$    against adversarial attacks?
				\vspace{-0.07cm}
	\end{minipage}}
	\vspace{-2mm}
\end{center}

\begin{wrapfigure}{r}{68mm}
\vspace{-5mm}
\centering{
\begin{tabular}{c}
\hspace*{-4mm}
\includegraphics[width=0.5\textwidth]{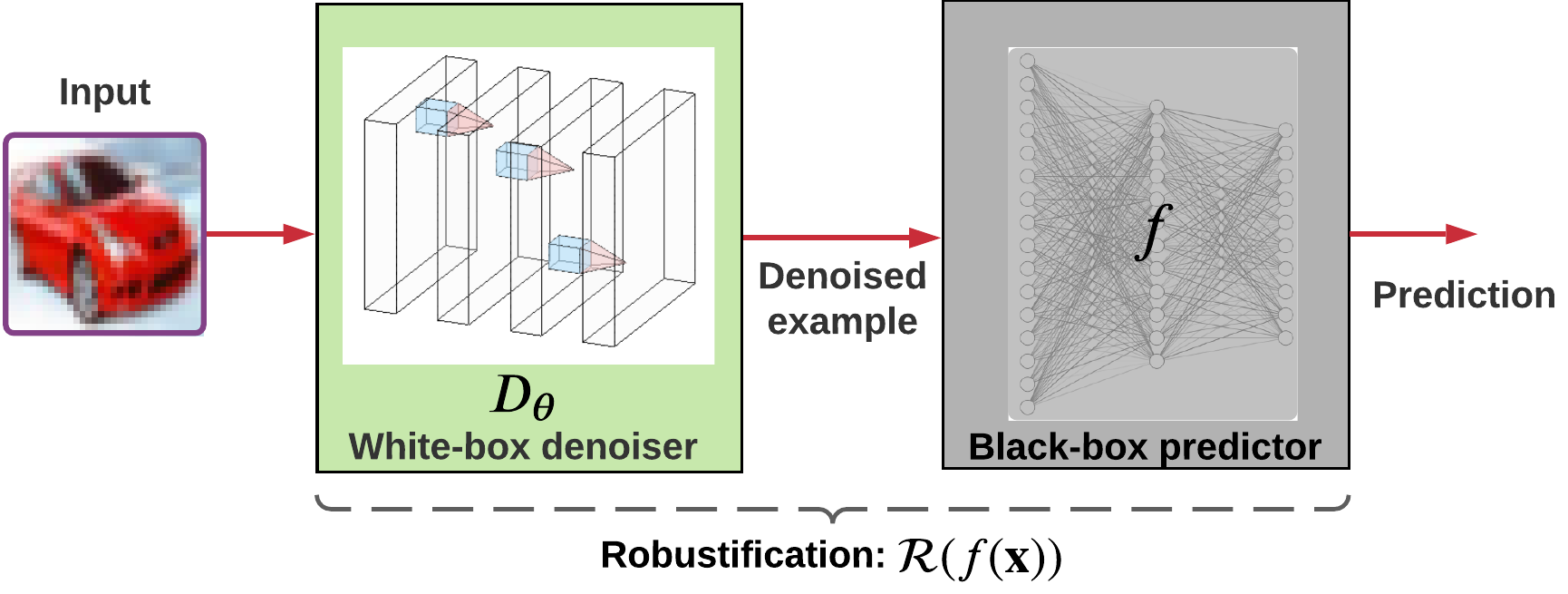}
\end{tabular}
}
\vspace{-5mm}
\caption{\footnotesize{{\DS}-based black-box defense.  
}}
\vspace{-1mm}
\label{fig: RS_black_box}
\end{wrapfigure}

\paragraph{Defensive operation.}
We next   provide a concrete formulation of the defensive operation $\mathcal R$.
In the literature, two principled defensive operations were  used: ($\mathcal R_1$) end-to-end AT \citep{madry2017towards,zhang2019theoretically,cohen2019certified}, and ($\mathcal R_2$) prepending a  defensive component  to a   base model \citep{meng2017magnet,salman2020denoised,aldahdooh2021adversarial}. 
The former ($\mathcal R_1$)  has achieved the state-of-the-art  robustness performance \citep{athalye2018obfuscated,croce2020reliable} but is not applicable to  black-box defense. 
By contrast, the latter ($\mathcal R_2$) is more compatible with black-box models. 
For example, \textit{denonised smoothing} ({\DS}), a recently-developed   $\mathcal R_2$-type approach  \citep{salman2020denoised}, gives a certified defense by   prepending a custom-trained denoiser to the  targeted model.
In this work,  we choose    {\DS} as the backbone of our defensive operation (Fig.\,\ref{fig: RS_black_box}).

In {\DS}, a  denoiser is integrated with a base model $f$ so that the augmented system becomes resilient to    Gaussian noise and thus plays a  role similar to  the {\RS}-based  certified defense \citep{cohen2019certified}.
That is, {\DS} yields  
\begin{align}
   \mathcal R ( f(\mathbf x) ) \Def f  (D_{\boldsymbol \theta} (\mathbf x ) ), 
   \label{eq: DS_model}
\end{align}
where 
$D_{\boldsymbol \theta}$ denotes the learnable denoiser (with   parameters $\boldsymbol \theta$) prepended to the (black-box) predictor $f$. Once  $D_{\btheta}$ is learned, then
the {\DS}-based smooth classifier, 
$
\argmax_{c} \mathbb P_{\boldsymbol \delta \in \mathcal N(\mathbf 0, \sigma^2 \mathbf I)}[  \mathcal R ( f(\mathbf x+\bdelta) ) = c ]
$, can achieve certified  robustness, where $c$ is a class label, $\boldsymbol \delta \in \mathcal N(\mathbf 0, \sigma^2 \mathbf I)$  denotes the standard Gaussian noise with variance $\sigma^2$, and 
$\argmax_{c} \mathbb P_{\boldsymbol \delta \in \mathcal N(\mathbf 0, \sigma^2 \mathbf I)}[ f (\mathbf x + \boldsymbol \delta)  = c ]$ signifies a smooth version  of $f$.

Based on \eqref{eq: DS_model}, the goal of black-box defense  becomes  to find the optimal denoiser $D_{\boldsymbol \theta}$ so as to achieve   satisfactory accuracy as well as adversarial  robustness. 
In the {\FO} learning paradigm, 
\citet{salman2020denoised} proposed a stability regularized denoising loss to train  $D_{\boldsymbol \theta}$: 
\begin{align}
  \hspace*{-2.5mm}  \begin{array}{ll}
    \displaystyle \minimize_{\boldsymbol \theta }     &   \hspace*{-1.5mm} 
  \mathbb E_{\bdelta \in \mathcal N(\mathbf 0, \sigma^2 \mathbf I), \mathbf x \in \mathcal{U}}    \underbrace{ \| D_{\btheta} (\mathbf x + \bdelta) - \mathbf x \|_2^2 }_\text{$\Def \ell_{\mathrm{Denoise}}(\btheta)$}  + \gamma \mathbb E_{\bdelta, \mathbf x} \underbrace{  \ell_{\mathrm{CE}} (\mathcal R ( f(\mathbf x+\bdelta) ), f    (\mathbf x)   ) }_\text{$\Def \ell_{\mathrm{Stab}}(\btheta)$},
    \end{array}
    \label{eq: DS}
\end{align}
where $\mathcal U$ denotes the training dataset, the first objective term $\ell_{\mathrm{Denoise}} (\btheta)$ corresponds to the  mean squared error (MSE) of image denoising,   the   
second objective term $\ell_{\mathrm{Stab}} (\btheta)$ measures the prediction stability  
through the cross-entropy (CE) between the outputs of the denoised input and the original input, and 
$\gamma > 0$ is a regularization parameter that strikes a balance between $\ell_{\mathrm{Denoise}} $ and $\ell_{\mathrm{Stab}}$.

We remark that problem \eqref{eq: DS} can be solved using the {\FO} gradient descent method if 
the base model $f$ is fully disclosed to the defender. However,
the black-box nature of $f$ makes the   gradients of the stability loss $\ell_{\mathrm{Stab}}(\btheta)$  infeasible to obtain. Thus,   we will develop a \textit{gradient-free} {\DS}-oriented defense.
\section{Method: A Scalable Zeroth-Order Optimization Solution}
\label{sec: method}

In this section, we begin by presenting a brief background on {\ZO} optimization, and elaborate on
the challenge of black-box defense  in high dimensions. 
Next,  we propose a novel ZO optimization-based {\DS}
 method that can  not only improve  model query complexity but also lead to certified robustness.

 \paragraph{ZO optimization.}
 In ZO optimization, the {\FO} gradient of a  black-box function $\ell(\mathbf w)$  (with a $d$-dimension variable $\mathbf w$)
 is approximated by  the   difference of two function values along  a set of random  direction vectors. This leads to the   randomized gradient estimate ({\RGE}) \citep{liu2020primer}:
 \begin{align}
     \hat \nabla_{\mathbf w}  \ell (\mathbf w) =  \frac{1}{q} \sum_{i=1}^q \left [ \frac{d}{\mu}\left (
     \ell (\mathbf w + \mu \mathbf u_i ) - \ell(\mathbf w)
     \right ) \mathbf u_i \right ] ,
     \label{eq: ZO_grad_est}
 \end{align}
where $\{ \mathbf u_i \}_{i=1}^q$ are $q$  random vectors drawn independently and uniformly    from the sphere of a unit ball, and $\mu > 0$ is a given small step
size, known as the smoothing parameter.
The rationale behind  \eqref{eq: ZO_grad_est} is that it provides an \textit{unbiased} estimate of the {\FO} gradient of the   Gaussian smoothing version of $\ell$ \citep{gao2018information}, with  variance in the order of $O(\frac{d}{q})$ \citep{liu2020primer}.
Thus, a large-scale problem (with large $d$)
yields
a large variance of  {\RGE} \eqref{eq: ZO_grad_est}. To reduce the variance, a large number of querying directions (\textit{i.e.}, $q$) is  then needed, with the worst-case query complexity in the order of $O(d)$. 
If $q = d$, then the least estimation variance can be achieved by the     coordinatewise  gradient estimate ({\CGE}) \citep{lian2016comprehensive,liu2018_NIPS}:
 \begin{align}
     \hat \nabla_{\mathbf w}  \ell (\mathbf w) =  \sum_{i=1}^d \left [ \frac{\ell (\mathbf w + \mu \mathbf e_i ) - \ell(\mathbf w)}{\mu} \mathbf e_i \right ],
     \label{eq: ZO_grad_est_coord}
 \end{align}
where $\mathbf e_i \in \mathbb R^d$
 denotes the $i$th elementary basis vector,
with $1$ at the $i$th coordinate and $0$s elsewhere.
For any off-the-shelf {\FO} optimizers, \textit{e.g.}, stochastic gradient descent (SGD), if we replace the {\FO} gradient estimate with the ZO gradient estimate, then we obtain the ZO counterpart of a {\FO} solver, \textit{e.g.}, ZO-SGD \citep{ghadimi2013stochastic}.

\mycomment{
\begin{figure}[htb]
\centering{
\begin{tabular}{c}
\hspace*{-3mm}
\includegraphics[width=0.6\textwidth]{Figures/Test_acc_motivation.png}
\end{tabular}
}
\vspace{-3mm}
\caption{\footnotesize{Performance of {\DS}-based black-box defense using a naive  ZO optimization approach. This ZO-DS method leverages the random gradient estimator (RGE) with query number $q = 192$.
}}
\vspace{-3mm}
\label{fig: RS_ZO_direct}
\end{figure}
}

\paragraph{Warm-up: A direct application of ZO optimization.}

A straightforward method to achieve the {\DS}-based black-box defense is to   solve problem \eqref{eq: DS} using ZO optimization directly. However, it will give rise to the difficulty of \textit{ZO optimization in high dimensions}.
Specifically, {\DS} requires to calculate the gradient of the defensive operation \eqref{eq: DS_model}. With the aid of ZO gradient estimation, we obtain 
\begin{align}
   \nabla_{\btheta} \mathcal R ( f(\mathbf x) ) = \frac{d D_{\btheta}(\mathbf x )}{d \btheta} \frac{d f(\mathbf z)}{d \mathbf z}\left. \right |_{\mathbf z = D_{\btheta}(\mathbf x )} \approx \frac{d D_{\btheta}(\mathbf x )}{d \btheta}  \hat{\nabla}_{\mathbf z} f(\mathbf z) \left. \right |_{\mathbf z = D_{\btheta}(\mathbf x )},
   \label{eq: ZO_direct_est}
\end{align}
 where with an abuse of notation, let 
 $d$ denote the dimension of $\mathbf x$ (yielding  
$D_{\btheta}(\mathbf x) \in \mathbb R^d$ and $\mathbf z \in \mathbb R^d$) and $d_\theta$ denote the dimension of $\btheta$,
 $\frac{d  D_{\btheta}(\mathbf x )}{d \btheta} \in \mathbb R^{d_\theta \times d}$ is the  
 Jacobian  matrix of the vector-valued function
 $ D_{\btheta}(\mathbf x )$, and
 $\hat{\nabla}_{\mathbf z} f(\mathbf z)$ denotes  the ZO 
 gradient estimate of $f$, 
 following \eqref{eq: ZO_grad_est} or \eqref{eq: ZO_grad_est_coord}.
 Since the   dimension  of an input    is typically large for image classification (\textit{e.g.}, $d = 3072$ for a CIFAR-10
 \begin{wraptable}{r}{61mm}
\vspace*{-2mm}
\centering
\resizebox{0.43\textwidth}{!}{
\begin{tabular}{c|c|c} 
\toprule
\hline
 Method & \begin{tabular}[c]{@{}c@{}}
Certified robustness (\%) \\
($\ell_2$ radius: $\epsilon = 0.5$)
\end{tabular} &  Standard accuracy (\%) \\
\hline
{\FODS} & 30.22 & 71.80 \\
\hline
\begin{tabular}[c]{@{}c@{}}
{\ZODS} \\
({\RGE}, $q = 192$)
\end{tabular}& 5.06 (\textcolor{blue}{$\downarrow$ 25.16}) & 44.81 (\textcolor{blue}{$\downarrow$ 26.99}) \\
\hline
\bottomrule
\end{tabular}}
\vspace*{-2mm}
\caption{\footnotesize{Performance comparison between {\FODS} \citep{salman2020denoised} and its direct ZO implementation {\ZODS} on (CIFAR-10, ResNet-110).}}
\label{tablle: ZO_DS_motivation}
\vspace*{-2mm}
\end{wraptable}
 image), it imposes \textit{two challenges}: (a) The variance of {\RGE} \eqref{eq: ZO_grad_est} will be  ultra-large  if the query complexity stays low, \textit{i.e.},   a small query number $q$ is used; And (b) the variance-least {\CGE}  \eqref{eq: ZO_grad_est_coord} becomes impracticable due to the need of ultra-high querying cost (\textit{i.e.}, $q = d$).
 Indeed,  Table\,\ref{tablle: ZO_DS_motivation} shows that the direct application of   \eqref{eq: ZO_direct_est}  into the existing {\FO}-{\DS} solver \citep{salman2020denoised}, which we call {\ZODS},
 yields over $25\%$ degradation in both standard accuracy and certified robustness evaluated at input
perturbations with $\ell_2$ norm less than $128/255$, where   pixels of an input image are normalized to $[\mathbf 0, \mathbf 1]$. We refer readers to Sec.\,\ref{Performance_Evaluations} for more details.

  \begin{wrapfigure}{r}{77mm}
\vspace{-3mm}
\centering{
\begin{tabular}{c}
\hspace*{-5mm}
\includegraphics[width=0.56\textwidth]{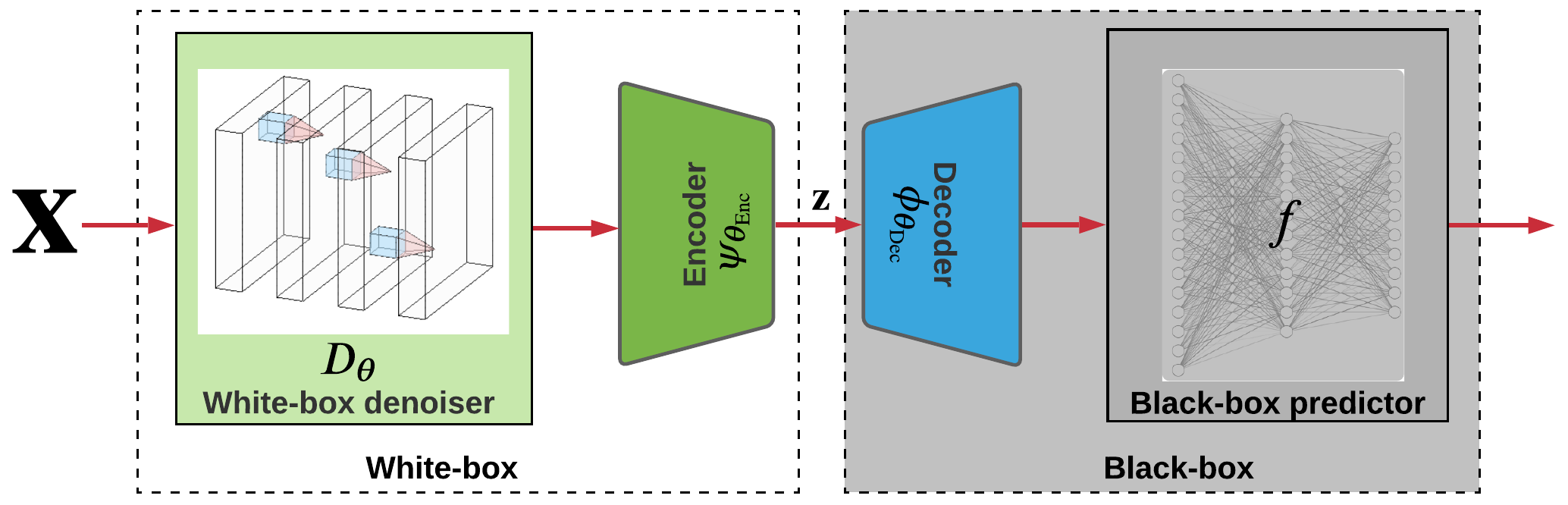}
\end{tabular}
}
\vspace{-3mm}
\caption{\footnotesize{Model architecture for {\ZOAEDS}.
}}
\vspace{-3mm}
\label{fig: AE_DS}
\end{wrapfigure}
\paragraph{ZO autoencoder-based {\DS} (ZO-AE-DS): A scalable solution to black-box defense.}
The   difficulty of ZO optimization in high dimensions prevents us from developing an effective {\DS}-oriented provable defense for black-box ML models. To tackle such problem, 
we introduce an Autoencoder (AE) to connect the  front-end
denoiser $D_{\btheta}$  with the back-end black-box predictive model  $f$ so that ZO   optimization can be conducted in a (low-dimension) feature embedding space. 
To be concrete, let 
$ \psi_{\boldsymbol \theta_{\mathrm{Dec}} } \circ \phi_{\boldsymbol \theta_{\mathrm{Enc}} }$ denote  {\AEnc}   consisting of the encoder (Enc) $\psi_{\boldsymbol \theta_{\mathrm{Enc}}}$ and the decoder (Dec) $\phi_{\boldsymbol \theta_{\mathrm{Dec}}}$, where $\circ$ denotes the function composition operation.
Plugging $\phi_{\boldsymbol \theta_{\mathrm{AE}}}$ between the  
denoiser $D_{\btheta}$  and the black-box predictor  $f$, we then extend the defensive operation \eqref{eq: DS_model} to the following (see Fig.\,\ref{fig: AE_DS} for  illustration):
\begin{align}
    \mathcal R_{\mathrm{new}}(f(\mathbf x)) \Def  \underbrace{ f \left ( \phi_{\boldsymbol \theta_{\mathrm{Dec}}} \left ( \mathbf z  \right )  \right ) }_\text{new black box}, ~~~~~~ \mathbf z = \underbrace{ \psi_{\boldsymbol \theta_{\mathrm{Enc}}}  \left ( D_{\btheta} (\mathbf x)  \right ) }_\text{new white box} ,
\label{eq: AE_DS}
\end{align}
where $\mathbf z \in \mathbb R^{d_z}$ denotes the low-dimension feature embedding  with $d_z < d$.
In \eqref{eq: AE_DS}, we integrate the decoder $\psi_{\boldsymbol \theta_{\mathrm{Dec}} }$ with the black-box predictor $f$ to construct a \textit{new} black-box model $f^\prime (\mathbf z) \Def f(\psi_{\boldsymbol \theta_{\mathrm{Dec}} }(\mathbf z))$, which enables us to derive a ZO gradient estimate of  \textit{reduced dimension}:
\begin{align}
  \nabla_{\btheta}  \mathcal R_{\mathrm{new}} ( f(\mathbf x) )
   \approx \frac{d  \phi_{\boldsymbol \theta_{\mathrm{Enc}}}   ( D_{\btheta} (\mathbf x)   ) }{d \btheta} \hat{\nabla}_{\mathbf z} f^\prime  (  \mathbf z     )  \left. \right |_{\mathbf z = \phi_{\boldsymbol \theta_{\mathrm{Enc}}}   ( D_{\btheta} (\mathbf x)   ) }.
   \label{eq: ZO_est_AE_DS}
\end{align}
Assisted by {\AEnc},  {\RGE} of  $\hat{\nabla}_{\mathbf z} f^\prime$   has a reduced variance from $O(
\frac{d}{q})$ to $O(
\frac{d_z}{q})$. Meanwhile,  the least-variance {\CGE} \eqref{eq: ZO_grad_est_coord} also becomes feasible by setting the query number as $ q = d_z$. 

Note that the eventual ZO estimate  \eqref{eq: ZO_est_AE_DS} is  a function of 
  the $d_\theta \times d_z $ Jacobian matrix 
$
\nabla_{\btheta}  [ \phi_{\boldsymbol \theta_{\mathrm{Enc}}}   ( D_{\btheta} (\mathbf x)   ) ] 
$. For ease of storing and computing the   Jacobian matrix, we derive  the following computationally-light alternative   of \eqref{eq: ZO_est_AE_DS} 
 (see derivation in Appendix\,{\ref{app: 1}}):
\begin{align}
  &  \nabla_{\btheta} \ell_{\mathrm{Stab}}(\btheta)
  \approx \nabla_{\btheta}
    [ \mathbf a^\top  \phi_{\boldsymbol \theta_{\mathrm{Enc}}}   ( D_{\btheta} (\mathbf x + \bdelta)   ) ], ~~
    \mathbf a =  \hat{\nabla}_{\mathbf z}
    \ell_{\mathrm{CE}} (f^\prime (\mathbf z), f(\mathbf x))
   \left. \right |_{\mathbf z = \phi_{\boldsymbol \theta_{\mathrm{Enc}}}   ( D_{\btheta} (\mathbf x + \bdelta)   ) }, \label{eq: grad_est_a}
\end{align}
where recall that $f^\prime (\mathbf z) = f(\psi_{\boldsymbol \theta_{\mathrm{Dec}} }(\mathbf z))$, and $\hat{\nabla}$ denotes the ZO gradient estimate given by  \eqref{eq: ZO_grad_est} or \eqref{eq: ZO_grad_est_coord}.
The computation advantage of   \eqref{eq: grad_est_a} is that the derivative operation $\nabla_{\btheta}$ can be applied to a scalar-valued inner product 
built upon a pre-calculated ZO gradient estimate $\mathbf a$.

\paragraph{Training {\ZOAEDS}.}
Recall from Fig.\,\ref{fig: AE_DS} that the proposed defensive system involves three components: denoiser $D_{\btheta}$, {\AEnc} $ \psi_{\boldsymbol \theta_{\mathrm{Dec}} } \circ \phi_{\boldsymbol \theta_{\mathrm{Enc}} }$, and pre-defined black-box predictor $f$. 
Thus, the parameters to be optimized include $\btheta$, $\btheta_{\mathrm{Dec}}$ and  $\boldsymbol \theta_{\mathrm{Enc}}$.
To train   {\ZOAEDS}, we adopt a two-stage training protocol. \ding{172} \textit{White-box pre-training on   {\AEnc}}: At the first stage, we pre-train 
the {\AEnc} model by calling a standard {\FO} optimizer (\textit{e.g.}, Adam) to minimize the    reconstruction loss 
$\mathbb E_{\mathbf x} \| \phi_{\btheta_{\mathrm{Dec}}} (\psi_{\btheta_{\mathrm{Enc}}} ( \mathbf x ) ) - \mathbf x \|_2^2$.  
The resulting {\AEnc} will be used as the initialization of the second-stage training. 
We remark that the  denoising model $D_{\btheta}$ can also be pre-trained. However, such a pre-training could hamper 
optimization, \textit{i.e.}, making the second-stage training over $\btheta$   easily trapped at a poor local optima. 
\ding{173} \textit{End-to-end   training}:
At the  second stage, 
we keep the pre-trained decoder  
$\phi_{\btheta_{\mathrm{Dec}}}$ intact and merge it into the black-box system as shown in Fig.\,\ref{fig: AE_DS}. 
We then optimize   $\btheta$  and  $\boldsymbol \theta_{\mathrm{Enc}}$  by minimizing the {\DS}-based training loss \eqref{eq: DS}, where the denoiser $D_{\btheta}$ and the defensive operation $\mathcal R$  are replaced by 
$\psi_{\btheta_{\mathrm{Enc}}} \circ D_{\btheta}$ and $\mathcal R_{\mathrm{new}}$ \eqref{eq: AE_DS}, respectively. In  \eqref{eq: DS},
  minimization over the stability loss  $\ell_{\mathrm{Stab}}(\btheta)$ 
  calls  the ZO   estimate of $\nabla_{\btheta}\ell_{\mathrm{Stab}}(\btheta)$, given by
  \eqref{eq: ZO_est_AE_DS}. In Appendix \,\ref{app: train_scheme}, different training schemes are discussed.
  
  \mycomment{
  However, such an estimate   relies on the computation of the $d_\theta \times d_z $ Jacobian matrix 
$
\nabla_{\btheta}  [ \phi_{\boldsymbol \theta_{\mathrm{Enc}}}   ( D_{\btheta} (\mathbf x)   ) ] 
$. To avoid storing and computing the   Jacobian matrix, we derive  an alternative form of 
 the ZO   estimate that is easy to implement (see derivation in Appendix\,\SL{\ref{app: 1}})
\begin{align}
  &  \nabla_{\btheta} \ell_{\mathrm{Stab}}(\btheta)
  \approx \nabla_{\btheta}
    [ \mathbf a^\top  \phi_{\boldsymbol \theta_{\mathrm{Enc}}}   ( D_{\btheta} (\mathbf x + \bdelta)   ) ], ~~
    \mathbf a =  \hat{\nabla}_{\mathbf z}
    \ell_{\mathrm{CE}} (f^\prime (\mathbf z), f(\mathbf x))
   \left. \right |_{\mathbf z = \phi_{\boldsymbol \theta_{\mathrm{Enc}}}   ( D_{\btheta} (\mathbf x + \bdelta)   ) }, \label{eq: grad_est_a}
\end{align}
where recall that $\psi_{\boldsymbol \theta_{\mathrm{Dec}} }$, and $\hat{\nabla}$ denotes the ZO gradient estimate given by  \eqref{eq: ZO_grad_est} or \eqref{eq: ZO_grad_est_coord}.
The computation advantage of   \eqref{eq: grad_est_a} is that the derivative operation $\nabla_{\btheta}$ can be applied to a scaled-valued inner product 
built upon a pre-calculated ZO gradient estimate $\mathbf a$.
}
\section{Experiments}
\label{Performance_Evaluations}

In this section, we demonstrate the effectiveness of our proposal
through extensive experiments. We will show that the proposed {\ZOAEDS} outperforms a series of baselines when robustifying black-box neural networks for secure    image classification and   image reconstruction.

\subsection{Experiment setup}

\paragraph{Datasets and model architectures.}

In the task of image classification, we focus on CIFAR-10 and STL-10 datasets. In   Appendix \,\ref{app: imagenet}, we demonstrate the effectiveness of ZO-AE-DS on the high-dimension ImageNet images.
In the task of image reconstruction, we consider the MNIST dataset. 
To build {\ZOAEDS} and its variants and baselines, we specify the prepended denoiser $D_{\btheta}$ as DnCNN \citep{zhang2017beyond}. 
We then implement task-specific {\AEnc} for different datasets. 
Superficially, the dimension of encoded feature embedding, namely, $d_{z}$ in \eqref{eq: AE_DS},   is set as 192, 576 and 192 for CIFAR-10, STL-10 and MNIST, respectively. The architectures of AE
are configured following \citep{mao2016image}, and ablation study on the choice of AE is shown in Appendix \,\ref{app: ae_arch}.
To specify the black-box image classification model, we choose 
ResNet-110 
for CIFAR-10 following \citep{salman2020denoised}, and ResNet-18 for STL-10. It is worth noting that  STL-10 contains 500 labeled $96\times96$ training images, and the pre-trained ResNet-18 achieves 76.6\% test accuracy that matches to state-of-the-art performance. 
For image reconstruction, we adopt a reconstruction network  consisting of convolution, deconvolution and ReLU layers, following \citep{raj2020improving}.

\paragraph{Baselines.}
We will consider two \textit{variants} of our proposed   ZO-AE-DS: \textit{i)
{\ZOAEDS} using {\RGE}} \eqref{eq: ZO_grad_est}, \textit{ii)  
{\ZOAEDS}  using {\CGE}} \eqref{eq: ZO_grad_est_coord}. 
In addition, we will compare {\ZOAEDS} with 
\textit{i) \textit{FO-AE-DS}}, \textit{i.e.}, the first-order implementation of {\ZOAEDS}, \textit{ii) \textit{\FODS}}, which    developed  in \citep{salman2020denoised}, \textit{iii)} \textit{\RS}-based certified training, proposed in  \citep{cohen2019certified}, and \textit{iv) {\ZODS}}, \textit{i.e.},
the ZO implementation of {\FODS} using {\RGE}. 
Note that {\CGE} is not applicable to {\ZODS} due to the   obstacle of high dimensions. 
To our best   knowledge, {\ZODS} is the only  query-based black-box defense baseline that can be directly compared with {\ZOAEDS}.

\paragraph{Training  setup.}
We build the training pipeline of the proposed {\ZOAEDS} following `Training {\ZOAEDS}' in Sec.\,\ref{sec: method}. To optimize the  denoising model $D_{\btheta}$,  
we  will cover two training schemes:   training from scratch, and  pre-training \& fine-tuning. 
In the scenario of training from scratch, we use Adam optimizer with learning rate $10^{-3}$ to train the model for 200 epochs and then use SGD optimizer with learning rate $10^{-3}$ drop by a factor of $10$ at every 200 epoch, where
the total number of epochs is 600.
As will be evident later,   training from scratch over $D_{\btheta}$ leads to better performance of {\ZOAEDS}. 
In the scenario of 
  pre-training \& fine-tuning, we use Adam optimizer to pre-train the denoiser $D_{\btheta}$
  with the MSE loss $\ell_{\mathrm{Denoise}}$ in \eqref{eq: DS}
  for 90 epochs and fine-tune the denoiser with $\ell_{\mathrm{Stab}}$ for $200$ epochs with learning rate $10^{-5}$ drop by a factor of $10$ every 40 epochs.
   When implementing the baseline {\FODS}, we use the best training setup provided  by \citep{salman2020denoised}. When implementing {\ZODS}, we  reduce the initial learning rate to $10^{-4}$ for training from scratch and $10^{-6}$ for pre-training \& fine-tuning
   to stabilize the convergence of ZO optimization.
  Furthermore,
  we set the smoothing parameter $\mu = 0.005$ for {\RGE} and {\CGE}. And
  to achieve a smooth predictor,  we set the Gaussian smoothing noise as $\boldsymbol \delta \in \mathcal N(\mathbf 0, \sigma^2 \mathbf I)$ with $\sigma^2 = 0.25$. With the help of matrix operations and the parallel computing power of the GPU, we optimize the training time to an acceptable range. The averaged one-epoch training time on a single Nvidia RTX A6000 GPU is about $\sim$ 1min and $\sim$ 29min for FO-DS and our proposed ZO method, ZO-AE-DS (CGE, $q=192$), on the CIFAR-10 dataset.

\paragraph{Evaluation metrics.}
In the task of robust image classification,
the performance  will be evaluated at \underline{s}tandard test \underline{a}ccuracy (SA) and \underline{c}ertified \underline{a}ccuracy (CA).
Here CA is a provable robust  guarantee of the   Gaussian smoothing version of a predictive model. Let us take {\ZOAEDS} as an example, the resulting smooth image classifier is given by  $f_{\mathrm{smooth}} (\mathbf x)\Def \argmax_{c} \mathbb P_{\boldsymbol \delta \in \mathcal N(\mathbf 0, \sigma^2 \mathbf I)}[  \mathcal R_{\mathrm{new}} ( f(\mathbf x+\bdelta) ) = c ]$, where $\mathcal R_{\mathrm{new}}$ is given by \eqref{eq: AE_DS}. 
Further,    a \textit{certified radius}   of  $\ell_2$-norm perturbation ball with respect to an input example   can be calculated   following the {\RS} approach provided in \citep{cohen2019certified}.
As a result, CA at a given $\ell_2$-radius $r$ is   the percentage of the correctly classified data points whose certified radii are larger than $r$.
Note that if $r =0$, then CA reduces to SA.

\begin{table*}[htb]
\center
\resizebox{0.85\textwidth}{!}{
\begin{tabular}{c|ccc|ccc|cccc} 
\toprule
\hline
 & \multicolumn{3}{c|}{FO} & \multicolumn{3}{c|}{ZO-DS}& \multicolumn{4}{c}{ZO-AE-DS (Ours)}\\
\hline
  $\ell_2$-\text{radius} $r$ & RS & FO-DS & FO-AE-DS &
  \begin{tabular}[c]{@{}c@{}}
  $q=20$ \\
({\RGE})
\end{tabular}
  &   \begin{tabular}[c]{@{}c@{}}
  $q=100$ \\
({\RGE})
\end{tabular} &   \begin{tabular}[c]{@{}c@{}}
  $q=192$ \\
({\RGE})
\end{tabular} & 
  \begin{tabular}[c]{@{}c@{}}
  $q=20$ \\
({\RGE})
\end{tabular}
&   \begin{tabular}[c]{@{}c@{}}
  $q=100$ \\
({\RGE})
\end{tabular} &   \begin{tabular}[c]{@{}c@{}}
  $q=192$ \\
({\RGE})
\end{tabular}  &   \begin{tabular}[c]{@{}c@{}}
  $q=192$ \\
({\CGE})
\end{tabular} \\
  \hline
  0.00 (SA)
  &$\mathbf{76.44}$ & 71.80 & $75.97$ & 19.50& 41.38 & 44.81 & 42.72& 58.61& 63.13& $\mathbf{72.23}$\\
  0.25 
  & $\mathbf{60.64}$ &51.74 & $59.12$ & 3.89& 18.05 & 19.16 & 29.57& 40.96& 45.69& $\mathbf{54.87}$\\
  0.50
  & $\mathbf{41.19}$&30.22 & $38.50$ & 0.60& 4.78 & 5.06 & 17.85& 24.28& 27.84& $\mathbf{35.50}$\\
  0.75
  &$\mathbf{21.11}$& 11.87 & $18.18$ & 0.03& 0.32 & 0.30 & 8.52& 9.45& 10.89& $\mathbf{16.37}$\\
\hline
\bottomrule
\end{tabular}}
\caption{
\footnotesize{SA (standard accuracy, \%) and CA (certified accuracy, \%) versus different values of  $\ell_2$-radius $r$. Note that SA corresponds to the case of   $r = 0$. In both {\FO} and {\ZO} blocks, the best accuracies for each $\ell_2$-radius are highlighted in \textbf{bold}.
}}
\label{cf_all}
\end{table*}

\subsection{Experiment results on image classification}

\paragraph{Performance on CIFAR-10.}
In Table\,\ref{cf_all}, we present certified accuracies of {\ZOAEDS} and its variants/baselines 
versus different $\ell_2$-radii in the setup of (CIFAR-10, ResNet-110). 
Towards a comprehensive comparison, different 
{\RGE}-based 
variants of {\ZOAEDS} and {\ZODS} are demonstrated using    the query number $q \in \{ 20, 100,  192 \}$.
First, the comparison between   {\ZOAEDS} and {\ZODS} shows that  our proposal significantly outperforms  {\ZODS} ranging from the low query number $q=20$ to  the high query number $q =192$ when {\RGE} is applied. 
Second, we observe that the use of {\CGE} yields the best CA and  SA (corresponding to $r = 0$). The application of {\CGE} is benefited from {\AEnc}, which reduces the dimension from $d = 32\times 32 \times 3$
to $d_z = 192$. In particular, {\CGE}-based {\ZOAEDS} improves the case studied in Table\,\ref{tablle: ZO_DS_motivation} from $5.06\%$ to $35.5\%$ at  the $\ell_2$-radius  $r = 0.5$.
Third, although {\FOAEDS}   yields CA improvement over {\FODS} in the white-box context, the improvement  achieved by {\ZOAEDS} (vs. {\ZODS}) for black-box defense is much more significant. This implies that 
 the performance of black-box defense  relies on a proper solution (namely, {\ZOAEDS}) to tackle the challenge of ZO optimization in high dimensions. 
 Fourth, 
 RS   outperforms the ZO methods. This is not surprising since RS is a known white-box certifiably robust training approach.
In Appendix\,\ref{app: 2}, we   demonstrate the consistent effectiveness   of {\ZOAEDS} under different denoisier and classifiers.
\vspace{-3mm}

\paragraph{Performance on STL-10.}
 In Table\,\ref{tablle:combination_new}, we 
 evaluate the performance of {\ZOAEDS} for STL-10 image classification. For comparison, we also represent the performance of {\FODS}, {\FOAEDS}, and {\ZODS}. Similar to Table\,\ref{cf_all}, the improvement brought by
our proposal
 over {\ZODS} is evident, with at least $10\%$ SA/CA improvement across different $\ell_2$-radii. 
 
\begin{wraptable}{r}{90mm}
\vspace*{-3mm}
\centering
\resizebox{0.6\textwidth}{!}{
\begin{tabular}{c|cccc} 
\toprule
\hline
 & \multicolumn{4}{c}{STL-10}\\
\hline

  $\ell_2$-\text{radius} $r$ & FO-DS & FO-AE-DS & \makecell{{\ZODS} \\ ({\RGE}, $q=576$)
  } & \makecell{{\ZOAEDS} \\ ({\CGE}, $q = d_z=576$)
  }   \\
  \hline
  0.00 (SA)
  & 53.36 & 
  54.26
  &  38.60  & 45.67 \\
  0.25 
  & 35.83 & 43.99 & 21.50 & 35.78  \\
  0.50
  & 21.61 & 34.85 & 9.58  &  26.70  \\
  0.75
  & 9.86 & 25.56  & 3.29 &  17.91 \\
\hline
\bottomrule
\end{tabular}}
\vspace*{-2mm}
\caption{\footnotesize{CA (certified accuracy, \%) vs. different $\ell_2$-radii for image classification on STL-10.
}} \label{tablle:combination_new}
\vspace*{-3mm}
\end{wraptable}

 When comparing {\ZOAEDS} with {\FODS}, we observe that ours introduces a $7\%$ degradation in SA (at $r = 0$). This is different from CIFAR-10 classification. There might be two reasons for  the   degradation of SA in STL-10. First, the  size of a STL-10 image is $9\times$ larger than a CIFAR-10 image. Thus, the over-reduced feature dimension could hamper SA. In this example, we set $d_z = 576$, which is only $3 \times $ larger than $d_z = 192$ used for CIFAR-10 classification. Second, 
 the variance of ZO gradient estimates   has a larger effect on the performance of STL-10 than that of CIFAR-10, since the former 
 only contains $500$ labeled   images, leading to a challenging training task.
 Despite the degradation of SA,  {\ZOAEDS} outperforms {\FODS} in CA, especially when facing a large $\ell_2$-radius. This is consistent with Table\,\ref{cf_all}. The rationale   is that {\AEnc} can be regarded as an extra smoothing operation for the image classifier, and thus improves certified robustness over {\FODS}, even if the latter is designed in a white-box setup. If we compare {\ZOAEDS} with {\FOAEDS}, then the {\FO} approach leads to the best performance due to the high-accuracy of gradient estimates.

\mycomment{
\begin{table}[htb]
\centering
\resizebox{0.95\textwidth}{!}{
\begin{tabular}{c|cccc|ccc} 
\toprule
\hline
 & \multicolumn{4}{c|}{(STL-10, ResNet-18)} & \multicolumn{3}{c}{(CIFAR-10, Wide-DnCnn, Vgg-16)}\\
\hline

  $\ell_2$-\text{radius} $r$ & FO-DS & FO-AE-DS & \makecell{{\ZODS} 
  } & \makecell{{\ZOAEDS} 
  }  & {\FODS} & {\FOAEDS} & {\ZOAEDS} \\
  \hline
  0.00 
  & 53.36 &  
  54.26
  &  38.60  & 45.67 & 65.09 &
 71.54 &
  72.97 \\
  0.25 
  & 35.83 & 43.99 & 21.50 & 35.78 &  44.52
  & 52.70
  & 54.92\\
  0.50
  & 21.61 & 34.85 & 9.58  &  26.70 &  25.21 
  & 32.01
  & 34.20\\
  0.75
  & 9.86 & 25.56  & 3.29 &  17.91 & 9.29
  & 13.52
  & 15.70\\
\hline
\bottomrule
\end{tabular}}
\caption{\footnotesize{CA vs. different $\ell_2$-radii under different combinations of denoiser and classifer.
\SL{[This table contains incorrect numbers. Why ZOAEDS is better than FOAEDS? }
}} \label{tablle:combination_new}
\end{table}
}

\begin{wrapfigure}{r}{65mm}
    \vspace*{-5mm}
\centerline{
\includegraphics[width=.43\textwidth,height=!]{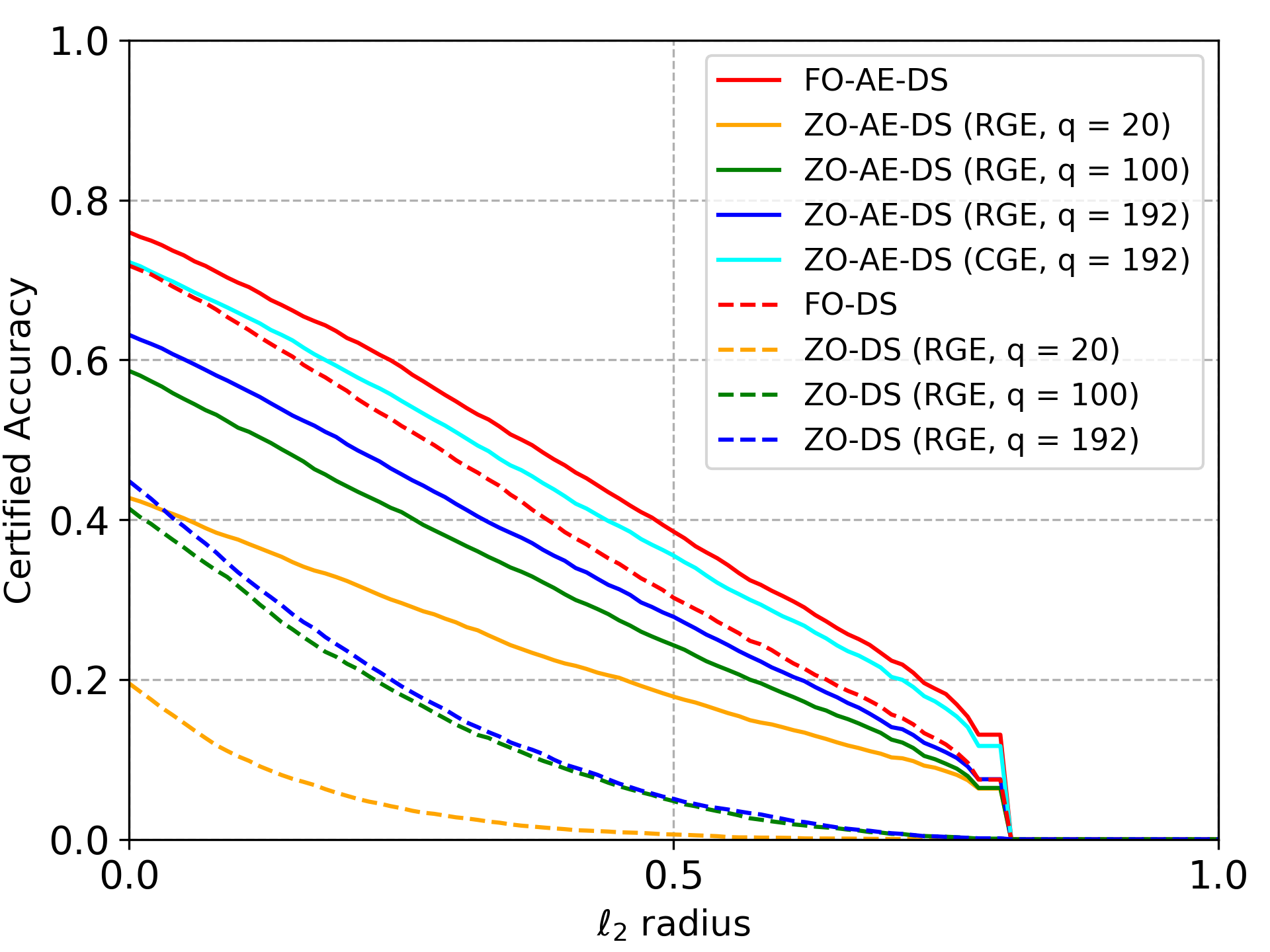}
}
\vspace*{-5mm}
\caption{\footnotesize{Comparison between 
 non-{\AEnc}-based and {\AEnc}-based methods in 
CA vs. different $\ell_2$-radii. Dashed lines: Models obtained by non-{\AEnc}-based methods; Solid lines: Models obtained by {\AEnc}-based methods. 
}}
  \label{fig:cf_all_overview}
  \vspace*{-5mm}
\end{wrapfigure}

\vspace*{-3mm}
\paragraph{Advantage of {\AEnc} on ZO optimization.}
Extended from Table\,\ref{cf_all},
 Fig.\,\ref{fig:cf_all_overview} presents the complete CA curve  of non-{\AEnc}-based  and
{\AEnc}-based methods vs. the value of $\ell_2$-radius in the example of (CIFAR-10, ResNet-110). As we can see, 
{\ZOAEDS} using {\RGE} with the smallest query number $q = 20$ has outperformed {\ZODS} using {\RGE} with the largest query number $q = 192$. This shows the vital role of {\AEnc} on ZO optimization. Meanwhile, consistent with Table\,\ref{cf_all} and Table\,\ref{tablle:combination_new}, the best model achieved by {\ZOAEDS} using {\CGE} could be even better than the {\FO} baseline {\FODS} since {\AEnc} could play a similar role on the smoothing operation. Furthermore,  as the query number $q$ increases,   the improvement of {\ZOAEDS} grows, towards the performance of {\FOAEDS}.

\begin{wraptable}{r}{70mm}
\vspace*{-5mm}
\centering
\resizebox{0.5\textwidth}{!}{
\begin{tabular}{c|c|c} 
\toprule
\hline
 & \multicolumn{2}{c}{\makecell{ {\ZOAEDS}  ({\CGE}, $q=192$)}}\\
\hline
  $\ell_2$-\text{radius} $r$ &  \makecell{Training from scratch}  &\makecell{Pre-training + fine-tuning} \\
  \hline
  0.00 &
72.23 & 59.74 \\
  0.25 
 & 54.87 & 42.61\\
  0.50
  &  35.50 & 26.26\\
  0.75
  & 16.37 & 11.13\\
\hline
\bottomrule
\end{tabular}}
\vspace*{-2mm}
\caption{\footnotesize{{\ZOAEDS} using  different denoiser training schemes under (CIFAR-10, ResNet-110).}} \label{tablle:training_scheme}
\vspace*{-5mm}
\end{wraptable}
\vspace*{-3mm}
\paragraph{Effect of   training scheme on {\ZOAEDS}.}
In Table\,\ref{tablle:training_scheme}, we present the impact of training scheme (over the denoiser   $D_{\btheta}$) on  the CA performance of {\ZOAEDS} versus different $\ell_2$-radii. 
Two training schemes, training from scratch and pre-training + fine-tuning, are considered. 
As we can see, training from scratch for $D_{\btheta}$ leads to the better performance of {\ZOAEDS} than pre-training + fine-tuning. This is because the application of pre-training to {$D_{\btheta}$} could make 
optimization easily get trapped at a local optima. We list other ablation studies in Appendix\,\ref{app: ablation}.

\subsection{Experiment results on image  reconstruction.}
In what follows, we apply the  proposed {\ZOAEDS}   to robustifying a black-box image reconstruction network. 
The goal of image reconstruction is to recover the original image from a noisy measurement. Following \citep{antun2020instabilities,raj2020improving}, we generate the noisy measurement following a linear observation model $\mathbf y = \mathbf A \mathbf x$, where $\mathbf A$ is a sub-sampling matrix (\textit{e.g.}, Gaussian sampling), and $\mathbf x$ is an original image. 
A pre-trained image reconstruction network \citep{raj2020improving} then takes $\mathbf A^{\top} \mathbf y$ as the input to recover $\mathbf x $. 
To evaluate the reconstruction performance, we adopt two metrics \citep{antun2020instabilities},  the root mean squared error (RMSE) and structural similarity (SSIM).  SSIM  is a supplementary metric to RMSE, since it gives an accuracy indicator when evaluating the similarity between the true image and its estimate at fine-level   regions. 
The vulnerability of image reconstruction networks to adversarial attacks, \textit{e.g.}, PGD attacks \citep{madry2017towards}, has been shown in  \citep{antun2020instabilities,raj2020improving,wolfmaking19}.

\begin{figure}[htbp]
\vspace*{-5mm}
\centering
\subfigure[\footnotesize{Ground truth}]{
\begin{minipage}[t]{0.33\linewidth}
\centering
\includegraphics[width=1.8in]{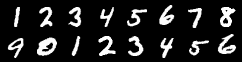}
\end{minipage}%
}%
\subfigure[\footnotesize{Standard}]{
\begin{minipage}[t]{0.33\linewidth}
\centering
\includegraphics[width=1.8in]{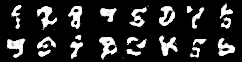}
\end{minipage}%
}%
\subfigure[\footnotesize{FO-DS}]{
\begin{minipage}[t]{0.33\linewidth}
\centering
\includegraphics[width=1.8in]{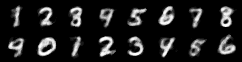}
\end{minipage}%
}%
\vspace*{-1mm}
\subfigure[\footnotesize{ZO-DS}]{
\begin{minipage}[t]{0.33\linewidth}
\centering
\includegraphics[width=1.8in]{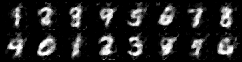}
\end{minipage}%
}%
\subfigure[\footnotesize{FO-AE-DS}]{
\begin{minipage}[t]{0.33\linewidth}
\centering
\includegraphics[width=1.8in]{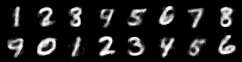}
\end{minipage}%
}%
\subfigure[\footnotesize{ZO-AE-DS}]{
\begin{minipage}[t]{0.33\linewidth}
\centering
\includegraphics[width=1.8in]{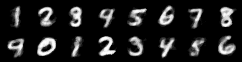}
\end{minipage}%
}%

\centering
\vspace*{-3mm}
\caption{\footnotesize{Visualization for Image Reconstruction under $\ell_2$ PGD attack (Step $= 40$, $\epsilon = 1.0$ ). Original: base reconstruction network. ZO-DS: RGE with $q=192$. ZO-AE-DS: CGE with $q=192$}}
\label{fig: recon_visual}
\end{figure}

When the image reconstructor is given as a black-box model, spurred by above, Table\,\ref{recon_mse_all} presents the 
performance  of image reconstruction using various training methods against adversarial attacks with different perturbation strengths. As we can see, 
compared to the normally trained image reconstructor (\textit{i.e.}, `Standard' in Table\,\ref{recon_mse_all}), all robustification methods   lead to degraded  standard image reconstruction performance in the non-adversarial context   (\textit{i.e.}, $\| \bdelta \|_2 = 0$). But the worst performance is provided by  {\ZODS}. 
When the perturbation strength increases, the model achieved by standard training becomes over-sensitive to adversarial perturbations, yielding the highest RMSE and the lowest SSIM. 
Furthermore, 
we   observe that the proposed black-box defense {\ZOAEDS}  yields very competitive and even better performance with respect to {\FO} defenses.  In Fig.\,\ref{fig: recon_visual}, we provide visualizations of the reconstructed images using different approaches at the presence of reconstruction-evasion PGD attacks. For example, the comparison between Fig.\,\ref{fig: recon_visual}-(f) and (b)/(d) clearly shows the robustness gained by {\ZOAEDS}.

\begin{table}[htb]
\vspace*{-3mm}
\center
\resizebox{0.7\textwidth}{!}{
\begin{tabular}{c|cc|cc|cc|cc|cc} 
\toprule
\hline
 \multicolumn{11}{c}{Image reconstruction on MNIST}\\
\hline
  \multirow{2}{*}{
    \begin{tabular}[c]{@{}c@{}}Method\end{tabular}
    } 
  &
  \multicolumn{2}{c|}{$\| \bdelta \|_2 = 0$} & \multicolumn{2}{c|}{$\| \bdelta \|_2 = 1$} & \multicolumn{2}{c|}{$\| \bdelta \|_2 = 2$} & \multicolumn{2}{c|}{$\| \bdelta \|_2 = 3$} & \multicolumn{2}{c}{$\| \bdelta \|_2 = 4$} \\
\cline{2-11}
  & RMSE & SSIM & RMSE & SSIM & RMSE & SSIM & RMSE & SSIM & RMSE & SSIM \\
  \hline
  \text{Standard}
    & 0.112 & 0.888 & 0.346 & 0.417 & 0.493 & 0.157 & 0.561 & 0.057 & 0.596 & 0.014\\
  \text{FO-DS} 
  & 0.143 & 0.781 & 0.168 & 0.703  & 0.221 & 0.544 & 0.278 & 0.417 & 0.331 & 0.337\\
  \text{ZO-DS}
  & 0.197 & 0.521 & 0.217 & 0.474  & 0.262  & 0.373 & 0.313 & 0.284 & 00.356 & 0.225\\
  \text{FO-AE-DS}
  &0.139 & 0.792 & 0.162 & 0.717  & 0.215  & 0.554 & 0.274 & 0.421 & 0.329 & 0.341\\
  \text{ZO-AE-DS} 
  & 0.141 & 0.79 & 0.164 & 0.718  & 0.217  & 0.551 & 0.277 & 0.42 & 0.33 & 0.339\\
\hline
\bottomrule
\end{tabular}}
\caption{
\footnotesize{Performance of image reconstruction using different methods at various attack scenarios. Here `standard' refers to the original image reconstructor without making any robustification. Four robustification methods are presented including {\FODS}, {\ZODS} ({\RGE}, $q=192$), {\FOAEDS}, and {\ZOAEDS} ({\CGE}, $q=192$). The performance metrics RMSE and SSIM are measured by adversarial example $(\mathbf x + \bdelta)$, generated by
  $40$-step $\ell_2$ PGD attacks under different  values of $\ell_2$ perturbation norm  $\|\boldsymbol \delta \|_2$.
}}\label{recon_mse_all}
\end{table}

\vspace{-3mm}
\section{Conclusion}

In this paper, we study the problem of black-box defense, aiming to secure black-box models against adversarial attacks using only input-output model queries. The proposed black-box learning paradigm is new to adversarial defense, but is also challenging to tackle because of the black-box optimization nature.  To solve this problem, we integrate denoised smoothing ({\DS}) with ZO (zeroth-order) optimization to build a feasible black-box defense framework. However, we find that the direct application of ZO optimization makes the defense ineffective and difficult to scale. We then propose {\ZOAEDS}, which leverages autoencoder ({\AEnc}) to bridge the gap between 
FO
and ZO optimization. We show that {\ZOAEDS} reduces the variance of ZO gradient estimates and improves the defense and optimization performance in a significant manner. Lastly, we evaluate the superiority of our proposal to a series of baselines in both image classification and image reconstruction tasks. 



\newpage
\section*{Acknowledgment}
Yimeng Zhang, Yuguang Yao, Jinghan Jia, and Sijia Liu are supported by the DARPA RED program.

{\small
\bibliographystyle{iclr2022_conference}
\bibliography{refs,ref_SL_adv,ref_SL_fair_self, ref_SL_ZO, ref_SL_AVs, ref-bi, ref_SL_pruning}
}

\newpage
\newpage
\clearpage

\setcounter{section}{0}


\setcounter{section}{0}
\setcounter{figure}{0}
\makeatletter 
\renewcommand{\thefigure}{A\@arabic\c@figure}
\makeatother
\setcounter{table}{0}
\renewcommand{\thetable}{A\arabic{table}}
\setcounter{mylemma}{0}
\renewcommand{\themylemma}{A\arabic{mylemma}}
\setcounter{algorithm}{0}
\renewcommand{\thealgorithm}{A\arabic{algorithm}}

\appendix
\section{Derivation of \eqref{eq: grad_est_a}}
\label{app: 1}
First, based on \eqref{eq: DS} and \eqref{eq: AE_DS}, the stability loss corresponding to {\ZOAEDS} is given by
\begin{align}\label{eq: app1_fs}
    \ell_{\mathrm{Stab}}(\btheta)
    =  \ell_{\mathrm{CE}} \left ( 
    f^\prime (\mathbf z) ,
    f    (\mathbf x)   \right ) \Def g(\mathbf z),  ~~\text{where}~~
    f^\prime (\mathbf z) = f \left ( \phi_{\boldsymbol \theta_{\mathrm{Dec}}} \left ( \mathbf z   \right )  \right )
    , ~~ \mathbf z =  \psi_{\boldsymbol \theta_{\mathrm{Enc}}}  \left ( D_{\btheta} (\mathbf x + \bdelta)  \right ).
\end{align}
We then take the derivative of $\ell_{\mathrm{Stab}}(\btheta)$ w.r.t. $\btheta$. This yields 
\begin{align}\label{eq: grad_appendix_1}
   \nabla_{\btheta} \ell_{\mathrm{Stab}}(\btheta)
    = \frac{d \mathbf z}{ d \btheta} \frac{d g(\mathbf z)}{d \mathbf z} \left . \right |_{\mathbf z = \psi_{\boldsymbol \theta_{\mathrm{Enc}}}  \left ( D_{\btheta} (\mathbf x + \bdelta)  \right )},
\end{align}
where 
    $\frac{d \mathbf z}{ d \btheta} \in \mathbb R^{d_\theta \times d}$ and $\frac{d g(\mathbf z)}{d \mathbf z} \in \mathbb R^{d}$.
    
    Since $g(\mathbf z)$ involves the black-box function $f$, we first compute its ZO gradient estimate following \eqref{eq: ZO_grad_est} or \eqref{eq: ZO_grad_est_coord} and obtain
    \begin{align}
       \frac{d g(\mathbf z)}{d \mathbf z} \left . \right |_{\mathbf z = \psi_{\boldsymbol \theta_{\mathrm{Enc}}}  \left ( D_{\btheta} (\mathbf x+ \bdelta)  \right )} \approx 
          \hat{\nabla}_{\mathbf z } g(\mathbf z) \left . \right |_{\mathbf z = \psi_{\boldsymbol \theta_{\mathrm{Enc}}}  \left ( D_{\btheta} (\mathbf x + \bdelta )  \right )}  \Def \mathbf a.
    \end{align}
    
    Substituting the above into  \eqref{eq: grad_appendix_1}, we obtain
    \begin{align}\label{eq: grad_appendix_2}
     \nabla_{\btheta} \ell_{\mathrm{Stab}}(\btheta)
    = \frac{d \mathbf z}{ d \btheta} \mathbf a = \begin{bmatrix}
    \frac{d \mathbf a^\top \mathbf z}{d \theta_1} \\ \frac{d \mathbf a^\top \mathbf z}{d \theta_2}\\ \vdots \\
    \frac{d \mathbf a^\top \mathbf z}{d \theta_{d_\theta}}
    \end{bmatrix} = \frac{d \mathbf a^\top \mathbf z }{d \btheta } = \nabla_{\btheta} [ \mathbf a^\top \phi_{\boldsymbol \theta_{\mathrm{Enc}}}   ( D_{\btheta} (\mathbf x + \bdelta )   ) ],
\end{align}
where the last equality holds based on \eqref{eq: app1_fs}.  This completes the derivation.

\section{Combination of Different Denoisers and Classifiers}
\label{app: 2}

Table\,\ref{tablle:vgg} presents the certified accuracies of our proposal using different denoiser models   (Wide-DnCnn vs. DnCnn) and   image classifier (Vgg-16).

\begin{table}[htb]
\centering
\resizebox{1\textwidth}{!}{
\begin{tabular}{c|ccc|ccc} 
\toprule
\hline
 & \multicolumn{3}{c|}{DnCnn \& VGG-16} & \multicolumn{3}{c}{Wide-DnCnn \& VGG-16} \\
\hline
  $\ell_2$-\text{radius} $r$ & FO-DS & FO-AE-DS & \makecell{ZO-AE-DS \\ ({\CGE}, $q = d_z=192$)
  } & FO-DS & FO-AE-DS & \makecell{ZO-AE-DS \\ 
  ({\CGE}, $q = d_z=192$)
  } \\
  \hline
  0.00 (SA)
  & 71.37 & 73.75 &  71.92  & 66.57 & 75.14 & 72.97 \\
  0.25 
  & 51.37 & 54.74 & 54.33 & 50.1 & 57.45  & 54.92\\
  0.50
  & 30.21 & 34.6 & 34.39  &  31.52 & 37.59 & 34.2 \\
  0.75
  & 11.72 & 15.45  & 15.36 &  13.94 & 17.64 & 15.7 \\
\hline
\bottomrule
\end{tabular}}
\caption{\footnotesize{CA (certified accuracy, \%) vs. different $\ell_2$-radii for different combinations of denoisers and classifier.
}} \label{tablle:vgg}
\end{table}


\section{Additional experiments and ablation studies}
\label{app: ablation}

In what follows, we will show the ablation study on the choice of AE architectures  in   Appendix\,\ref{app: ae_arch}.
Afterwards, we will show the performance of FO-AE-DS versus different training schemes in  Appendix\,\ref{app: train_scheme}.
Finally, we will show the performance of our proposal on the high-dimension ImageNet images   in   Appendix\,\ref{app: imagenet}.

\subsection{The performance of FO-AE-DS with different AutoEncoders.}
\label{app: ae_arch}

Table. \ref{tablle:ae_arch} presents the certified accuracy performance of FO-AE-DS with different autoencoders (AE).  As we can see, if AE-96 is used (namely, the encoded dimension is half of AE-192 used in the paper), then we observe a slight performance drop. This is a promising result as we can further reduce the query complexity by choosing a different autoencoder since the use of CGE has to be matched with the encoded dimension.


\begin{table}[htb]
\centering
\resizebox{0.3\textwidth}{!}{
\begin{tabular}{c|c|c} 
\toprule
\hline
  $\ell_2$-\text{radius} $r$ &  AE-96 & AE-192  \\
  \hline
  0.00 (SA)
  &  75.57&  $\mathbf{75.97}$   \\
  0.25 
  &  58.07 & $\mathbf{59.12}$ \\
  0.50
  &  37.09 & $\mathbf{38.50}$\\
  0.75
  &  17.05  & $\mathbf{18.18}$  \\
\hline
\bottomrule
\end{tabular}}
\caption{\footnotesize{CA (certified accuracy, \%) vs. different $\ell_2$-radii for FO-AE-DS with different AutoEncoders.
}} \label{tablle:ae_arch}
\end{table}

\subsection{The performance of FO-AE-DS with different training schemes.}
\label{app: train_scheme}

Table. \ref{tablle:train_scheme} presents the certified accuracy of FO-AE-DS (first-order implementation of ZO-AE-DS) with different training schemes. Training both denoiser and encoder is the default setting. As we can see, only training the denoiser would bring performance degradation, and training both denoiser and AE does boost the performance. It is worth noting that FO-AE-DS with "train the denoiser and AE" training scheme can be regarded as the FO-DS treating the combination of the original denoiser and the same AE used in FO-AE-DS as a new denoiser, which cannot be implemented for ZO-AE-DS since the decoder of ZO-AE-DS is merged into the black-box classifier and its parameters cannot be updated. Furthermore, the key of the introduced AE is to reduce the variable dimension for Zeroth-Order (ZO) gradient estimation.

\begin{table}[htb]
\centering
\resizebox{0.7\textwidth}{!}{
\begin{tabular}{c|c|c|c|c} 
\toprule
\hline
  $\ell_2$-\text{radius} $r$ & FO-DS & \makecell{FO-AE-DS \\ (only train \\ the denoiser)
  } & \makecell{FO-AE-DS \\ (train  the denoiser \\ and encoder)} & \makecell{FO-AE-DS \\ (train the denoiser \\ and the AE)}  \\
  \hline
  0.00 (SA)
  & 71.80 & 73.34 & 75.97  & 75.76  \\
  0.25 
  & 51.74 & 55.61 & 59.12 & 58.14  \\
  0.50
  & 30.22 & 35.68 & 38.50 &  38.88 \\
  0.75
  & 11.87 & 15.92  & 18.18 &  18.48   \\
\hline
\bottomrule
\end{tabular}}
\caption{\footnotesize{CA (certified accuracy, \%) vs. different $\ell_2$-radii for FO-AE-DS with different training schemes.
}} \label{tablle:train_scheme}
\end{table}

\subsection{The performance of ZO-AE-DS on ImageNet Images.}
\label{app: imagenet}

To evaluate the performance of ZO-AE-DS on the Restricted ImageNet (R-ImageNet) dataset, a 10-class subset of ImageNet with 38472 images for training and 1500 images for testing, similar to \citep{tsipras2018robustness}.
Due to our limited computing resources, we are not able to scale up our experiment to the full ImageNet dataset, but the purpose of evaluating on high-dimension images remains the same. In the implementation of ZO-AE-DS, we choose an AE with an aggressive compression (130:1), which is to compress the original $3 \times 224 \times 224$ images into the $1152 \times 1 \times 1$ feature dimension.
We compare the certified accuracy (CA) performance of our proposed ZO-AE-DS (using CGE) with the black-box baseline ZO-DS, and the white-box baselines FO-DS and FO-AE-DS. Results are summarized in the following table.

As we can see, (1) when considering the black-box classifier, the proposed ZO-AE-DS still significantly outperforms the direct ZO implementation of DS. This shows the importance of variance reduction of query-based gradient estimates. (2) Since ZO-AE-DS and FO-AE-DS used an aggressive AE structure, the performance drops compared to FO-DS. (3) the use of high-resolution images would make the black-box defense much more challenging. However, ZO-AE-DS is still a principled black-box defense method that can achieve reasonable performance.

\begin{table}[htb]
\centering
\resizebox{0.65\textwidth}{!}{
\begin{tabular}{c|c|c|c|c} 
\toprule
\hline
  $\ell_2$-\text{radius} $r$ & FO-DS & FO-AE-DS 
  & \makecell{ZO-AE-DS \\ (RGE, q =1152 \\ and encoder)} & \makecell{ZO-AE-DS \\ (CGE, q=1152)}  \\
  \hline
  0.00 (SA)
  & 89.33 & 71.07 & 26.93  & 63.60  \\
  0.25 
  & 81.67 & 63.40 & 18.40 & 52.80  \\
  0.50
  & 68.87 & 53.60 & 11.67 &  43.13 \\
  0.75
  & 49.80 & 42.87  & 5.53 &  32.73   \\
\hline
\bottomrule
\end{tabular}}
\caption{\footnotesize{CA (certified accuracy, \%) vs. different $\ell_2$-radii for FO-AE-DS on ImageNet Images.
}} \label{tablle:imagenet}
\end{table}

\end{document}